# A New Look at the Easy-Hard-Easy Pattern of Combinatorial Search Difficulty

**Dorothy L. Mammen**  MAMMEN@CS.UMASS.EDU
*Department of Computer Science*
*University of Massachusetts*
*Amherst, MA 01003, U.S.A.*

**Tad Hogg**  HOGG@PARC.XEROX.COM
*Xerox Palo Alto Research Center*
*3333 Coyote Hill Road*
*Palo Alto, CA 94304, U.S.A.*

## Abstract

The easy-hard-easy pattern in the difficulty of combinatorial search problems as constraints are added has been explained as due to a competition between the decrease in number of solutions and increased pruning. We test the generality of this explanation by examining one of its predictions: if the number of solutions is held fixed by the choice of problems, then increased pruning should lead to a monotonic decrease in search cost. Instead, we find the easy-hard-easy pattern in median search cost even when the number of solutions is held constant, for some search methods. This generalizes previous observations of this pattern and shows that the existing theory does not explain the full range of the peak in search cost. In these cases the pattern appears to be due to changes in the size of the minimal unsolvable subproblems, rather than changing numbers of solutions.

## 1. Introduction

Recently, many authors have shown that the solution cost for various kinds of combinatorial search problems follows a pattern of easy-hard-easy as a function of how tightly constrained the problems are. For example, this pattern appears for graph coloring as a function of the average graph connectivity (Cheeseman, Kanefsky, & Taylor, 1991; Hogg & Williams, 1994), for propositional satisfiability (SAT) as a function of the ratio of number of clauses to number of variables (Cheeseman et al., 1991; Mitchell, Selman, & Levesque, 1992; Crawford & Auton, 1993; Gent & Walsh, 1994b), and for constraint satisfaction problems (CSPs) as a function of the number of nogoods (Williams & Hogg, 1994) and constraint tightness (Smith, 1994; Prosser, 1996).

This regularity raises the possibility of determining, prior to search, the likely difficulty of problems. Unfortunately, this is not yet possible because of the high variance associated with the observations. This is compounded by the fact that a single problem can be viewed as belonging to a variety of problem classes, each with somewhat different transition points. Thus one important direction for improvement is to investigate whether there are simple additional parameters that can reduce this variance and allow predictions with higher confidence.

One approach to this question is based on the explanation of the easy-hard-easy pattern as a competition between changes in the number of solutions and pruning of unproductive search paths as a function of some measure of the degree to which the problems are constrained. In particular this predicts that problems with many solutions tend to be easier, on average, than those with fewer for a given number of constraints. Thus, at least one aspect of the high variance in search cost appears to be due to the variance in number of solutions in the problems of a fixed degree of constraint. This observation has motivated the introduction of additional parameters describing problem structure based on a more precise specification of the number of solutions (Hogg, 1996).

In this paper we investigate the generality of this explanation by examining problems for which the number of solutions is restricted, including cases where the number is specified exactly to be either zero or one. If the peak in search cost in fact arises generally from a competition between changes in the number of solutions and pruning, cases with a fixed number of solutions should not show a peak. However, we find that a peak continues to appear in these cases for some sophisticated search algorithms, while it fails to appear in other cases. This calls into question the generality of the explanation based on number of solutions, and also suggests that a search for additional problem structure parameters based solely on reducing the variance in the number of solutions is not likely to be sufficient to accurately predict search cost. However, some structural aspect of problems is likely to be involved. Specifically, we present data showing that the smallest of the problem's minimal unsolvable subproblems correlates well with search cost.

In the next section we describe some classes of search problems. We then review the pattern of search behavior and the current theoretical explanation for it. In the following section we uncover some limitations of this explanation by examining problems with some specification on their number of solutions. This shows the easy-hard-easy pattern is a more general phenomenon than suggested by current explanations. We then suggest an alternative explanation related to problem structure, and present data for unsolvable problems showing a positive relationship between this problem structure parameter, the minimum size of minimal unsolvable subproblem, and search cost. This same problem structure parameter may explain differences in search cost among solvable problems with equal numbers of solutions, as well. Finally, we discuss some of the implications of these observations and make suggestions for obtaining a better understanding and greater predictability for hard search problems.

## 2. Some Classes of Search Problems

In common with many previous studies of the transition phenomenon, we use random binary CSPs and graph coloring as example classes of search problems. This section describes how the problems were generated and searched.

### 2.1 Random CSPs

The constraint satisfaction problems used in most of our experimental results consist of 10 variables with three possible values for each one, and in some cases, we repeated experiments with problems of 20 variables. Problem constraints are specified by a number of binary nogoods, i.e., assignments to a pair of variables that are considered to be inconsistent. The search problem is then to find a consistent complete assignment, i.e., a value for each variable that does not include any of the inconsistent pairs.

We generated problems in a number of ways to fully sample the range of behaviors. In the first method ("generate-select") we generate CSPs by randomly selecting the specified number of binary nogoods. To produce classes of problems with restrictions on their number of solutions, we determine the number of solutions of these randomly generated problems and retain only those satisfying these restrictions. For example, to produce a class of solvable problems, only those with a solution are included. Similarly, to produce a class of problems with a fixed number of solutions, only those problems with exactly the specified number of solutions are retained.

This random generation method gives a simple, uniform selection from the various problem classes. However, it can also be very inefficient in generating problems. For instance, with few nogoods, most randomly generated problems are solvable, hence requiring a large number of random trials to obtain even a few unsolvable cases.

To address this problem, we also used more efficient ("hill-climbing") methods. Specifically, for generating solvable problems with many nogoods, starting with a randomly generated unsolvable problem, we removed constraints at random until the problem became solvable, then restored the number of constraints removed with constraints chosen randomly, but with the requirement that the problem not become unsolvable again.

For generating unsolvable problems with few nogoods, the hill-climbing method started with a randomly generated solvable problem, removed the constraint that constrained the problem the least (the one whose removal increased the number of solutions the least), and added a randomly chosen constraint that resulted in a problem with fewer solutions than the problem had before the constraint removal. If, having removed one constraint, no other constraint could decrease the number of solutions, the constraint that increased the number of solutions the least was chosen – a slightly backwards step. To speed this process up, we checked only one third of the possible constraints before giving up, choosing the one that increased the number of solutions the least, and starting another iteration.

Other methods for generating problems with specified requirements on the number of solutions have also been studied. One popular method for solvable problems is to randomly select an assignment to all of the variables (a pre-specified solution) and then, during the random selection of nogoods, avoid any that are inconsistent with this pre-specified solution. This tends to emphasize problems with many solutions and results in instances that are somewhat easier than uniform random selection. Cha & Iwama (1995) have also used the approach of generating problems with specific attributes, for SAT problems, using the AIM generators (Asahiro, Iwama, & Miyano, 1993).

We solved these problems using dynamic backtracking (Ginsberg, 1993) in most cases, using random variable and value ordering. For comparison, we also did some searches with simple chronological backtrack instead. The search cost is measured as the number of nodes explored.

### 2.2 Graph Coloring

We also experimented with the 3-coloring problem. This constraint satisfaction problem consists of a graph and the requirement to assign each node one of three colors so that no pair of nodes linked by an edge have the same color. Each edge in the graph defines some binary nogoods for the problem, namely all pairs of assignments giving the same color to the two nodes connected by the edge. Thus each edge in the graph gives three binary nogoods. A convenient measure of the number of constraints is $\gamma$, the connectivity or average degree of the nodes in the graph. This is equal to twice the number of edges in the graph divided by the number of nodes, because each edge is incident on two nodes. For the 100-node graphs we studied, the number of binary nogoods is given by $150\gamma$.

In this case, we used a simple chronological backtrack search in combination with the Brelaz heuristic for variable and value ordering (Johnson, Aragon, McGeoch, & Schevon, 1991). This heuristic assigns the most constrained nodes first (i.e., those with the most distinctly colored neighbors), breaking ties by choosing nodes with the most uncolored neighbors, and with any remaining ties broken randomly. The colors are considered in a fixed ordering for all of the nodes in the search. As a simple optimization, the search never changes the colors selected for the first two nodes. Any such changes would amount to unnecessarily repeating the search with a permutation of the colors for unsolvable cases. Search cost is measured by the number of nodes explored.

## 3. The Easy-Hard-Easy Pattern

In this section, we present an example of the how search cost varies with the tightness of constraints for a class of problems, and describe how this behavior can be understood in terms of changes in the structure of the problems, independent of particular search algorithms. This review and summary of previous studies of the transition then forms a basis for comparison with the new results presented in subsequent sections.

### 3.1 An Example

Figure 1 shows a typical example of the easy-hard-easy pattern as a function of the constrainedness of the problem. Problems with few or many constraints tend to be easy to solve while those with an intermediate number are more difficult. The fraction of solvable problems is also shown in Figure 1, scaled from 1.0 on the left to 0.0 on the right. This illustrates that the hard problems are concentrated in the so-called "mushy region" (Smith & Dyer, 1996) where the probability of a solution is changing from 1.0 to 0.0. In particular, the peak in search cost is near the "crossover point," the point at which half the problems are solvable and half unsolvable. For this problem class, the crossover point occurs at just over 75 binary nogoods, and the peak in dynamic backtracking solution cost occurs at about 85 binary nogoods.

In all of our results in this paper, we include 95% confidence intervals (Snedecor & Cochran, 1967). These intervals for the estimate of the median obtained from our samples are given approximately by the percentiles $50 \pm 100/\sqrt{N}$ of the data, where $N$ is the number of samples. For the estimate of fractions the intervals are given approximately by $f \pm 2\sqrt{f(1-f)/N}$, where $f$ is the estimated value of the fraction. Finally, for the estimate of means the intervals are approximately $\overline{x} \pm 1.96\sigma/\sqrt{N}$ where $\overline{x}$ is the estimate of the mean and $\sigma$ the standard deviation of the sample. In many cases in this paper, there are sufficient samples to make these intervals smaller than the size of the plotted points.

A key point from examples such as this is that the difficult instances within a class of search problems tend to be concentrated near a particular value of the constraint tightness (here measured by the number of binary nogoods). Because this behavior is seen for a variety of search methods, it indicates this concentration does not depend much on the details of the search algorithm. Instead, it appears to be associated with a change in the properties of the problems themselves, namely their solvability.

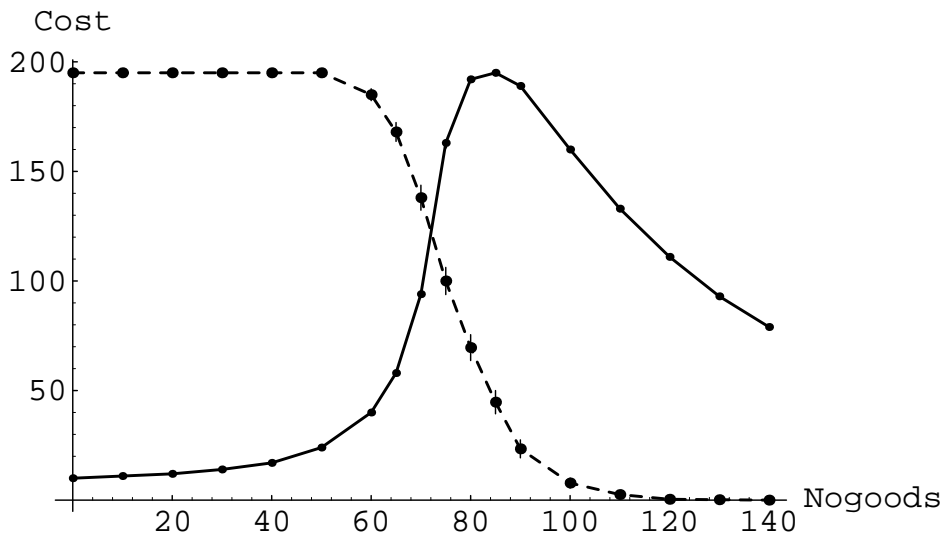

Figure 1: Typical transition pattern. Median solution cost for dynamic backtracking (solid line) and probability of a solution (dashed line) as a function of number of nogoods. Each point represents 1000 problems of 10 variables and domain size 3, each solved 100 times. Error bars showing 95% confidence intervals are included, but are in some cases smaller than the size of the plotted points.

### 3.2 An Explanation

These observations raise a number of questions, such as why a peak in search cost exists, why the peak occurs near the transition from mostly solvable to mostly unsolvable problems and is thus independent of the particular search algorithm, and why this behavior is seen for a large variety of constraint satisfaction problems.

The existing explanation for the concentration of hard problems relies on a competition between changes in the number of solutions and the amount of pruning provided by the problem constraints (Williams & Hogg, 1994). With few constraints, there are many solutions so the search is usually easy. As constraints are added the number of solutions drops rapidly, making problems harder. But the new constraints also increase the pruning of unproductive search choices, tending to make search easier. When there are few constraints, the decrease in the number of solutions overwhelms the increase in pruning, giving harder problems on average. Eventually the last solution is eliminated and all that remains is the increased pruning from additional constraints, leading to easier problems. Thus the phase transition, the point at which there is a precipitous change from solvability to unsolvability, more or less coincides with the peak in solution cost. All these effects become more pronounced as larger problems are considered, leading to sharper peaks and more abrupt transitions. This qualitative description explains many features of the observed behavior. This pruning explanation was also offered by Cheeseman et al. (1991) with respect to finding Hamiltonian circuits in highly constrained problems.

This explanation can also be used to obtain a quantitative understanding of the behavior. For instance, the location of the transition region can be understood by an approximate theory predicting that the cost peak occurs when the expected number of solutions equals one (Smith & Dyer, 1996; Williams & Hogg, 1994). In our example there are $3^{10}$ possible assignments to the 10 variables in the problem. There are $\binom{10}{2}3^2 = 405$ possible binary nogoods for the problem, which counts the number of ways to select a pair of variables and the different assignments for that pair. A given complete assignment for the 10 variables will be a solution provided each of the selected binary nogoods does not use the same assignment for its pair of variables as in the given complete assignment. This leaves $\binom{10}{2}(3^2 - 1) = 360$ possible choices for the binary nogoods. Thus the expected number of solutions is given by

$$3^{10} \times \frac{\binom{360}{m}}{\binom{405}{m}}$$

for problems with $m$ randomly selected binary nogoods. This expression equals one at $m = 82.9$, the location of the observed cost peak. Furthermore, because the expected number of solutions grows exponentially with the number of variables when $m$ is smaller than this threshold value and decreases exponentially to zero when $m$ is larger, the range of $m$ values over which the expected number of solutions is near one rapidly decreases as variables are added. This accounts for the observed sharpening of the transition for larger problems.

A further quantitative success of relating the search cost peak to transition phenomena is the evaluation of scaling behavior of the transition and search cost peak (Kirkpatrick & Selman, 1994; Gent, MacIntyer, Prosser, & Walsh, 1995).

## 4. Search Difficulty and Solvability

In this section we take a closer look at the behavior of the search cost, specifically, by examining how the behavior depends on whether the problem has a solution and, if so, the number of solutions.

### 4.1 Search Behavior

Figure 2 shows the median dynamic backtracking solution cost for solvable and unsolvable random CSPs generated as described above, for problems with number of variables $n = 10$ and $n = 20$, with domain size three. Except where specified otherwise in the figure caption, for problems of 10 variables we generated 1000 solvable and 1000 unsolvable problems for each point, and for problems of 20 variables we generated 500 solvable and 500 unsolvable problems for each point, using the "generate-select" method. We also generated unsolvable problems of 10 variables with 10 to 70 nogoods using the "hill-climbing" method. We overlap the range of problems generated by the two methods to show how the different generation methods affect search cost.

This figure clearly shows the easy-hard-easy pattern of solution cost for both solvable and unsolvable problems, for both problem sizes. The two methods of generating unsolvable problems give distinct curves: the unsolvable problems generated by the "hill-climbing" method are harder than those generated by the "generate-select" method. Nonetheless, both sets of problems show the same easy-hard-easy pattern.

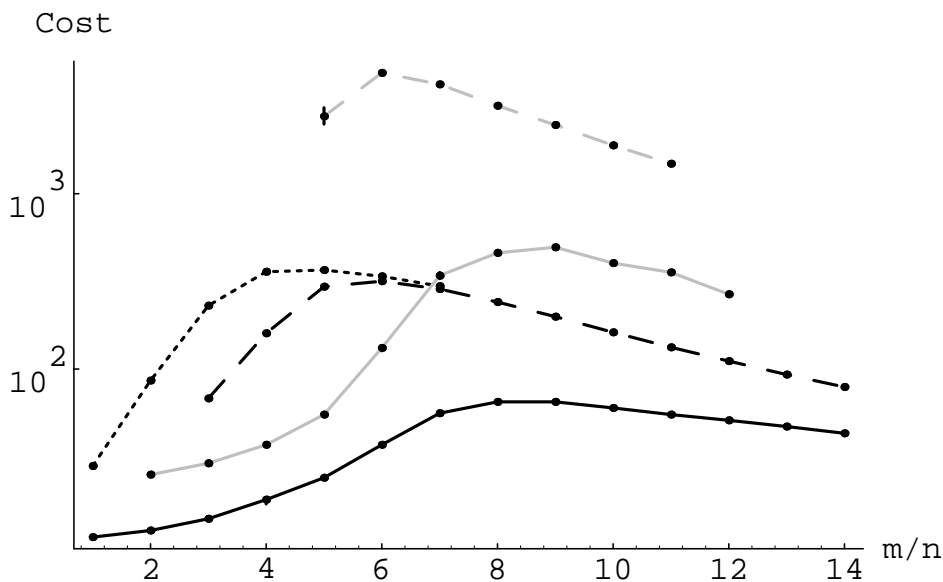

Figure 2: Median solution cost using dynamic backtracking for solvable (solid lines) and unsolvable (dashed and dotted lines) problems with number of variables $n = 10$ (black lines) and $n = 20$ (gray lines) as a function of number of nogoods divided by problem size, $m/n$. All problems were generated using the "generate-select" method except for the unsolvable problems shown by the dotted line, which were generated using the "hill-climbing" method. For problems of size 10, each point is the median of 1000 problems solved 100 times, except for unsolvable problems generated by "generate-select" at $m/n = 3$ (30 nogoods) and solvable problems at $m/n = 14$ (140 nogoods), which are based on 100 problems. For problems of size 20, each point is the median of 500 problems solved 100 times, except for unsolvable problems at $m/n = 5$ (100 nogoods) and solvable problems at $m/n = 12$ (240 nogoods), which are based on 15 and 35 problems, respectively. Error bars showing 95% confidence intervals are included, but in most cases are smaller than the size of the plotted points.

Another example with the same behavior is shown in Figure 3 for the median search cost for instances of 3-coloring of random graphs. In contrast to Figure 2, the solvable and unsolvable cases have similar median search costs near the peaks. This is because, as described above, the graph coloring searches for unsolvable cases used the symmetry with respect to permutations of the colors to avoid unnecessary search. Without this optimization, the costs for unsolvable cases would be six times greater than the values shown in the figure. Similar peaks are seen for other classes of graphs, such as connected ones, although at somewhat different values of $\gamma$.

These data show that both random CSPs and graph coloring problems exhibit an easy-hard-easy pattern for solvable and unsolvable problems considered separately.

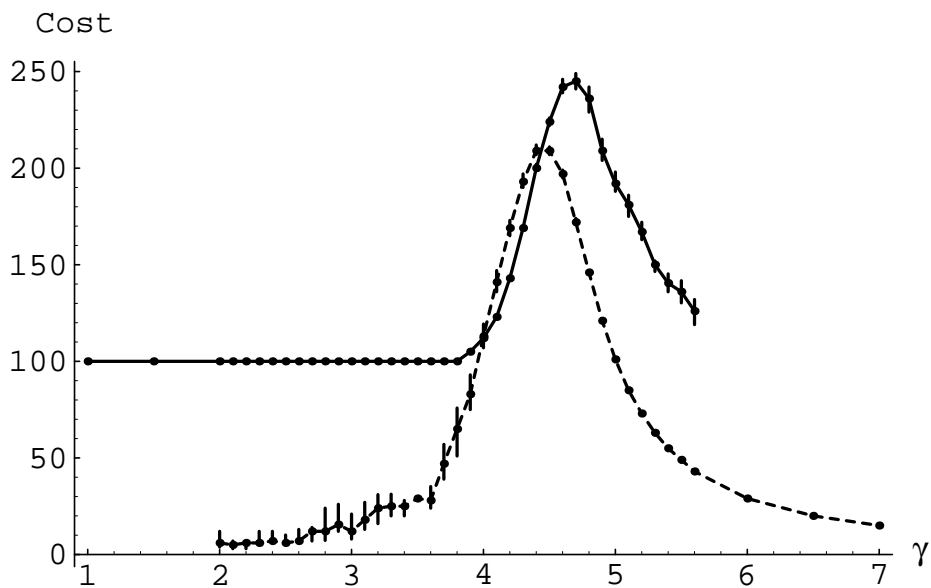

Figure 3: Median solution cost for 3-coloring random graphs with 100 nodes as a function of connectivity $\gamma$ using backtrack search with the Brelaz heuristic. The solid and dashed curves correspond to solvable and unsolvable cases respectively. These results started with 100,000 random graphs at each value of $\gamma$, and additional samples were generated at the extremes to produce at least 100 samples for each point. For random graphs, the crossover from mostly solvable to mostly unsolvable occurs around a connectivity of 4.5. Error bars showing 95% confidence intervals are included.

## 4.2 Solvable Problems

A peak in search cost for solvable problems such as we observed has also been seen extensively in studies of local-repair search methods and for problems generated with a pre-specified solution (Yugami, Ohta, & Hara, 1994; Kask & Dechter, 1995; Williams & Hogg, 1994). These search methods start with some assignment to all of the variables in the problem and then attempt to adjust them until a solution is found. Generally, such methods are not systematic searches: they can never determine that a problem has no solution. Thus empirical studies of these methods are restricted to consider solvable problems and incidentally provide a useful examination of the properties of solvable problems.

Furthermore, a study of satisfiability problems with backtracking search is consistent with a peak in cost for solvable problems (Mitchell et al., 1992), but there were insufficient highly constrained solvable problems to make a definite conclusion for the behavior with many constraints.

How does the existence of a peak for solvable problems fit with the explanation given above? Certainly an explanation based on a transition from solvable to unsolvable problems cannot apply directly to the class of solvable problems. However, the competition between increased pruning and decreased number of solutions still applies. As shown in Figure 4, the number of solutions for solvable random CSPs of size 10 at first decreases rapidly as constraints are added but then nears its minimum value of one, giving a slower decrease.

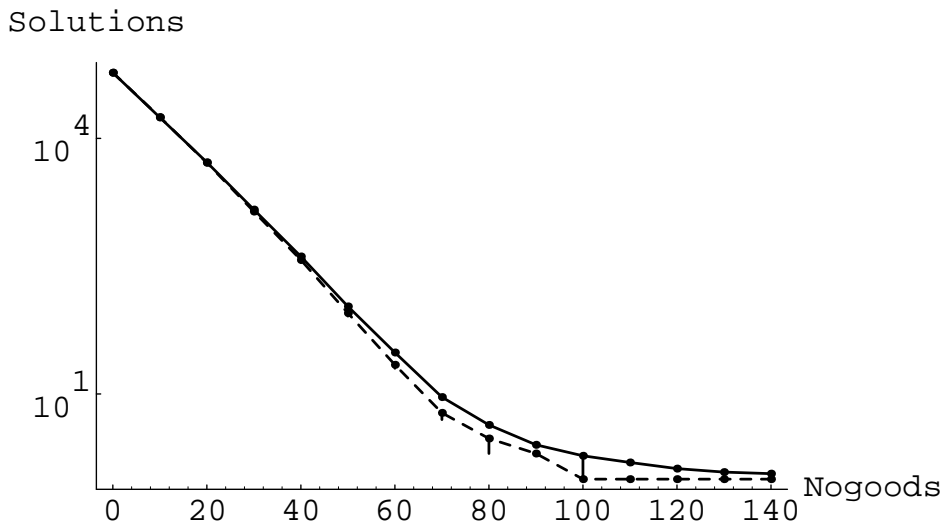

Figure 4: Mean (solid) and median (dashed) number of solutions on a log scale as a function of the number of binary nogoods, for solvable problems with 10 variables, 3 values each, based on 1000 problems generated by the "generate-select" method at each multiple of 10 binary nogoods, except for 140 nogoods, which is based on 100 problems. At 0 nogoods there are $3^{10} = 59049$ solutions. Error bars showing 95% confidence intervals are included.

Except for the change in minimum value from 0 to 1 solution, this behavior for the number of solutions is qualitatively similar to that for the general case including both solvable and unsolvable problems. The additional constraints continue to increase the pruning of unproductive search paths. Thus the explanation given above might continue to apply but now predicts the peak will be at the point where solutions can drop no further (i.e., one solution) rather than becoming unsolvable (i.e., zero solutions).

Figure 5 evaluates this idea. This figure shows how the fraction of problems with at least two solutions changes as a function of the number of nogoods divided by the problem size for random CSPs with 10 and 20 variables. For problems of size 10, the second to last solution disappears, on average, between 90 and 100 nogoods: the median number of solutions has dropped to 2 by 90 nogoods, and to 1 by 100 nogoods (Figure 4). The peak in solution cost for solvable problems is slightly lower than this, at between 80 and 90 nogoods, close to the crossover point of Figure 5 where half the solvable problems have only one solution. This is perhaps close enough to be consistent with the explanation given above. However, this relationship does not hold for problems of size 20. For this class of problems, the cost peak of solvable problems is at around 180 nogoods ($m/n = 9$), whereas the point at which half the problems have just one solution has still not been reached by 240 nogoods ($m/n = 12$). At 180 nogoods, the median number of solutions is 4 (mean is 10.0), and at 240 nogoods, the median is still 2 (mean is 1.83). This is inconsistent with the explanation that the cost peak for solvable problems is due to the increasing effect of pruning given no possible further decrease in number of solutions.

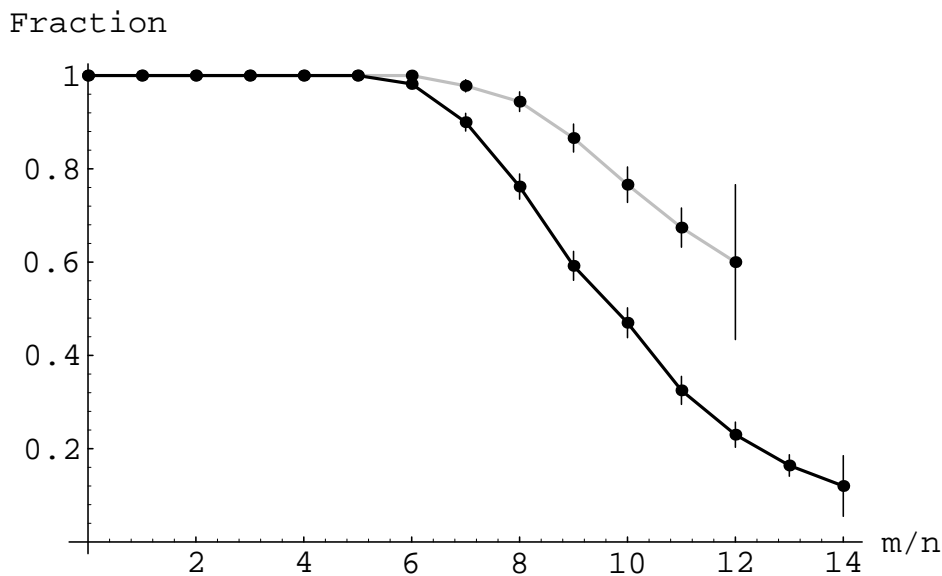

Figure 5: Fraction of problems with at least two solutions as a function of number of nogoods divided by problem size, for problems of size 10 (black line) and size 20 (gray line). Data for problems of size 10 are based on 1000 solvable problems created by the "generate-select" method at each point, except for 100 solvable problems at $m/n = 14$ (140 nogoods). Data for problems of size 20 are based on 500 solvable problems at each point, except for 20 solvable problems at $m/n = 12$ (240 nogoods), also created by the "generate-select" method. Error bars showing 95% confidence intervals are included.

Since the explanation depending on a change to insolubility does not apply, and the pruning versus number of solutions explanation does not fit the data, some other factors must be at work to produce the easy-hard-easy pattern for solvable problems. We suspect the explanation is related to the idea of minimal unsolvable subproblems. A minimal unsolvable subproblem is a subproblem that is unsolvable, but for which any subset of variables and their associated constraints is solvable; Gent & Walsh (1996) have referred to this aspect of SAT problems as the minimal unsatisfiable subset. The idea is that once a few bad choices have been made initially, such that the remainder of the problem becomes unsolvable, unsolvability is much harder to determine for some problems than for others. In particular, the more variables that are involved in a minimal unsolvable subproblem, the harder it is to determine that the subproblem is unsolvable. We make the conjecture that the cost peak for solvable problems is tied to the average size of the minimal unsolvable subproblem once a choice has been made that results in the remaining problem being unsolvable.

### 4.3 Problems With a Fixed Number of Solutions

A more interesting case is the behavior of the problems with no solutions shown in Figures 2 and 3. As a further example, Figure 6 shows the solution cost for problems with exactly one solution. This also shows a peak. These observations on problems with zero or one solution show that even with the number of solutions held constant, problems exhibit an easy-hard-easy pattern of solution cost.

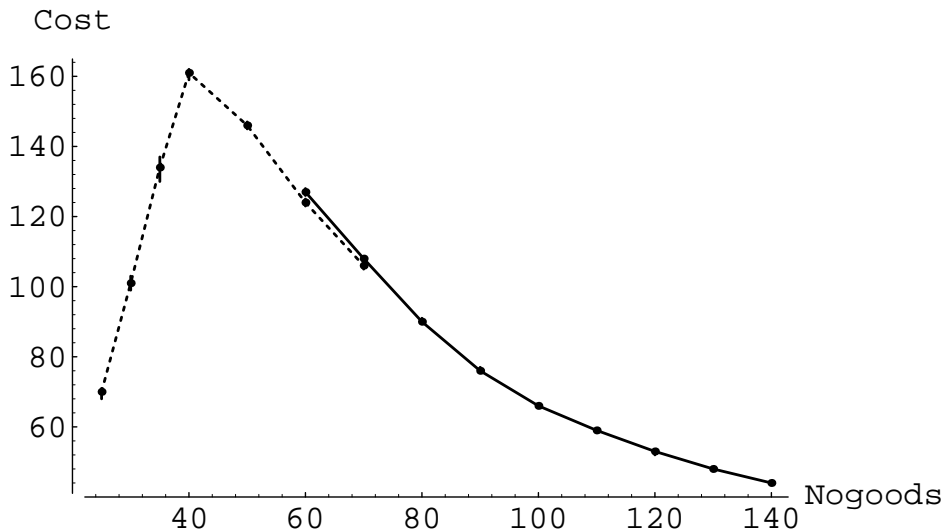

Figure 6: Median solution cost as a function of number of nogoods for problems of 10 variables, 3 values each, with exactly one solution, generated using the "generate-select" method (solid line), and by hill-climbing down to one solution starting from solvable problems with many solutions produced using "generate-select" (dotted line), solved using dynamic backtracking. Each point is the median of 1000 problems each solved 100 times, except for hill-climbing generated problems at 25, 30 and 35 nogoods and "generate-select" generated problems at 140 nogoods, of which there are 100. Error bars showing 95% confidence intervals are included.

According to the explanation of the transition, if the number of solutions is held constant then the increase in pruning will be the only factor, giving rise to a monotonic decrease in search cost as constraints are added. Instead, we see in Figures 2, 3 and 6 that even when the number of solutions is held fixed at zero or one, there is still a peak in solution cost, and at a smaller number of nogoods. Thus the existing explanation does not capture the full range of behaviors. Instead, it appears that there are other factors at work in producing hard problems. By focusing more closely on these factors we can hope to gain a better understanding of the structure of hard problems, which may lead to more precise predictions of search cost.

We also investigated the effect of algorithm on the pattern of solution cost in unsolvable problems by repeating the search of random CSPs using chronological backtrack. A comparison of chronological backtracking search with our previous dynamic backtrack search results for unsolvable problems is shown in Figure 7. In this figure, the curves for dynamic backtracking are the same as those for the unsolvable problems shown in Figure 2, except that here the cost curves are shown on a logarithmic scale. Interestingly, we do not see a peak in search cost for unsolvable problems using the less sophisticated method of chronological backtrack.

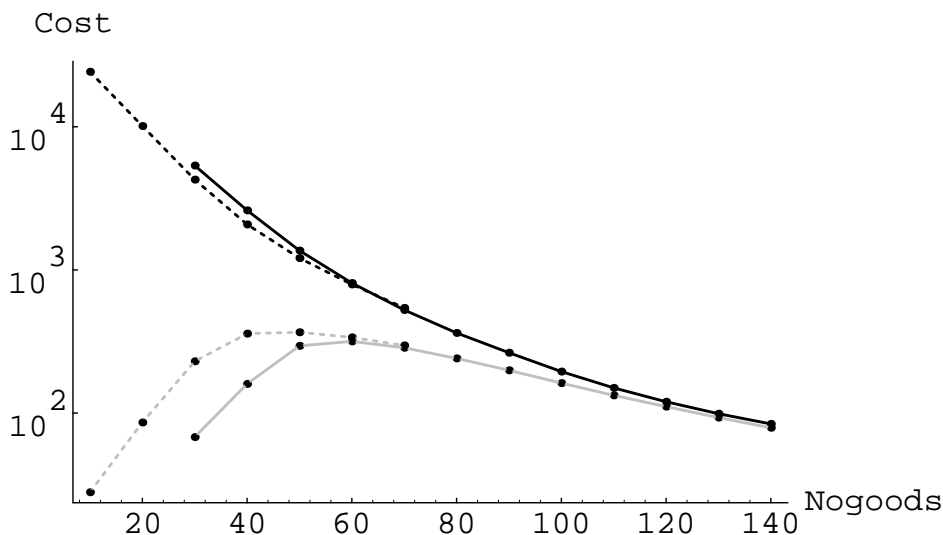

Figure 7: Comparison of median solution cost on a log scale using the same sets of unsolvable problems for chronological backtracking (black) and dynamic backtracking (gray). Dotted lines are for problems generated using the "hill-climbing" method, solid lines for the "generate-select" method. Each point is the median of 1000 problems each solved 100 times, except for the "generate-select" method at 30 nogoods, which is based on 100 problems. Error bars showing 95% confidence intervals are included, but are smaller than the size of the plotted points.

This observation raises an important point: the easy-hard-easy pattern is not a universal feature of search algorithms for problems restricted to a fixed number of solutions. This suggests that the competition between number of solutions and pruning, when it occurs, is sufficiently powerful to affect most search algorithms (very simple methods, such as generate-and-test, do not make use of pruning and show a monotonic increase in search cost as the number of solutions decreases), but that only some algorithms are able to exploit the features of weakly constrained problems with a fixed number of solutions that make them easy.

In contrast to our observations, a monotonic decrease in cost has been reported for unsolvable binary random constraint problems (Smith & Dyer, 1996) and for unsolvable 3SAT problems (Mitchell et al., 1992). In the case of 3SAT, the explanation may well be choice of algorithm. Indeed, Bayardo & Schrag (1996) recently found that incorporating conflict-directed backjumping and learning into the Tableau algorithm made a difference of many orders of magnitude in problem difficulty specifically for rare, "exceptionally hard," unsatisfiable problems in the underconstrained region. It would be interesting to see whether the easy-hard-easy pattern for unsolvable problems would appear using their algorithm.

With respect to Smith & Dyer's (1996) observations, the difference may be due to the range of problems generated, resulting from different problem generation methods. Smith and Dyer used a method akin to our "random" generation method, that is, generating problems without regard for solvability, then separating out the unsolvable ones. With this method, the "hit rate" for unsolvable problems in the underconstrained region is very low. It is possible that Smith and Dyer's data do not extend down to the point at which the cost of unsolvable problems begins to decrease simply because they stopped finding unsolvable problems before that point.

There are two possible reasons why we might have found unsolvable problems using random generation further into the underconstrained region, where Smith and Dyer did not. One possibility is that since we were specifically interested in unsolvable problems as far into the underconstrained region as possible, we may have spent more computational effort generating in that region. Indeed, at 40 nogoods, unsolvable problems occurred with frequency $4.5 \times 10^{-5}$, and at 30 nogoods, with frequency $7.75 \times 10^{-7}$. At that rate, problems at 30 nogoods took about six hours apiece to generate.

A second possibility relates to the details of the generation methods. In Smith and Dyer's random generation method, every pair of variables had exactly the same number of inconsistent value pairs between them. This implies a degree of homogeneity in the distribution of the nogoods. On the other hand, in our random generation method, each variable-value pair had an equal probability of being selected as a nogood, independent of one another. Thus it was at least possible in our generation method, though still of low likelihood, for the nogoods to occasionally clump, and to produce an unsolvable problem. This idea is discussed further in Section 5.

The difference in our observation and Smith & Dyer's (1996) reinforces an important point: that a relatively subtle difference in generation methods can affect the class of problems generated. While the nogoods will be more or less evenly distributed on average using our generation method, they will also be clumped with some probability, whereas with Smith and Dyer's generation method, a homogeneous distribution over variable pairs is guaranteed. These types of problems could be different enough to sometimes produce different behavior.

## 5. Minimal Unsolvable Subproblems

Our observations on classes of problems with restrictions on the number of solutions they may have shows that the common identification of the peak in solution cost with the algorithm-independent transition in solvability seen in general problem classes does not capture the full generality of the easy-hard-easy pattern.

For solvable problems, this explanation could be readily modified to use a transition in the existence of solutions beyond those specified by the construction of the class of problems and symmetries those problems might have that constrain the allowable range of solutions. This modification is a simple generalization of the existing explanation based on the competition between the number of solutions and pruning. However, our data for solvable problems do not support this explanation, in that the search cost peak and disappearance of the second to last solution coincide only roughly for $n = 10$, and not at all for $n = 20$.

Furthermore, when the number of solutions is held constant, competition between increased pruning and decreasing number of solutions cannot possibly be responsible for a peak in solution cost. The decrease in search cost for highly constrained problems (to the right of the peak) is adequately explained by the prevailing explanation, based on the increase in pruning with additional constraints. But this does not explain why weakly constrained problems are also found to be easy, at least for some search methods. The low cost of unsolvable problems in the underconstrained region is a new and unexpected observation in light of previous studies of the easy-hard-easy pattern and its explanation. This raises the question of whether there is a different aspect of problem structure that can account for the peak in search cost for problems with a fixed number of solutions.

One possibility that is often mentioned in this context is the notion of critically constrained problems. These are problems just on the boundary between solvable and unsolvable problems, i.e., neither underconstrained (with many solutions) nor overconstrained (with none). This notion forms the basis for another common interpretation of the cost peak. That is, these critically constrained problems will typically be hard to search (because most of the constraints must be instantiated before any unproductive search paths can be identified) and, since they are concentrated at the transition (Smith & Dyer, 1996), give rise to the search peak. This explanation does not include any discussion of the changes in pruning capability as constraints are added. Taken at face value, this explanation would predict no peak at all for solvable problems or when the number of solutions is held constant, because such classes have no transition from solvable to unsolvable problems. Moreover, this description of critically constrained problems is not simply a characteristic of an individual problem but rather is partly dependent on the class of problems under consideration because the exact location of the transition depends on the method by which problems are generated. This observation makes it difficult to characterize the degree to which an individual problem is critically constrained purely in terms of structural characteristics of that problem. By contrast, a measure such as the number of solutions is well defined for individual problem instances, which facilitates using its average behavior for various classes of problems to approximately locate the transition region. Thus, as currently described, the notion of critically constrained problems does not explain our observations nor does it give an explicit way to characterize individual problems.

A more precisely defined alternative characteristic is the size of minimal unsolvable subproblems. As we mentioned in Section 4.2, a minimal unsolvable subproblem is a subproblem that is unsolvable, but for which any subset of variables and their associated constraints is solvable.

Some problems have more than one minimal unsolvable subproblem. For example, a problem might have one minimal unsolvable subproblem of five variables, and another, different one, of say, six. We computed *all* minimal unsatisfiable subproblems for all of the 10-variable unsolvable problems we had generated. We found a monotonic positive relationship between mean number of minimal unsolvable subproblems and number of nogoods. For example, problems with 140 nogoods have an average of 35 minimal unsolvable subproblems (range 4 to 64, standard deviation 8.7); those with 90 nogoods have about six (range 1 to 23, standard deviation 3.6); and problems with 50 or fewer nogoods rarely have more than one minimal unsolvable subproblem. Similarly, Gent & Walsh (1996) observed that unsatisfiable problems in the underconstrained region tend to have small and unique minimal unsatisfiable subsets.

The behavior of the size of the smallest minimal unsolvable subproblem as a function of the number of nogoods is shown in Figure 8. Comparing with Figure 2, we see that the peak in the minimum size of minimal unsolvable subproblems matches the location of the search cost peak for unsolvable problems. This result is independent of whether we plot the smallest minimal unsolvable subproblem size, as shown in Figure 8, or medians or means, which we have not shown here. Moreover, the location of the peaks in minimal unsolvable subproblem size for the different generation methods correspond to the location of their respective search cost peaks. The peak in both search cost and minimal unsolvable subproblem size occurs at around 40 nogoods for problems generated using the "hill-climbing" method, and significantly higher, around 60 nogoods, for problems generated using the "generate-select" method. The strong correspondence between minimal unsolvable subproblem size and search cost is very suggestive that minimal unsolvable subproblem size is a structural characteristic of problems that plays an important role in search cost. By contrast, number of minimal unsolvable subproblems does not match the pattern of search cost. As mentioned above, it increases monotonically with number of nogoods, suggesting that it does not play a primary role in explaining search cost for unsolvable problems.

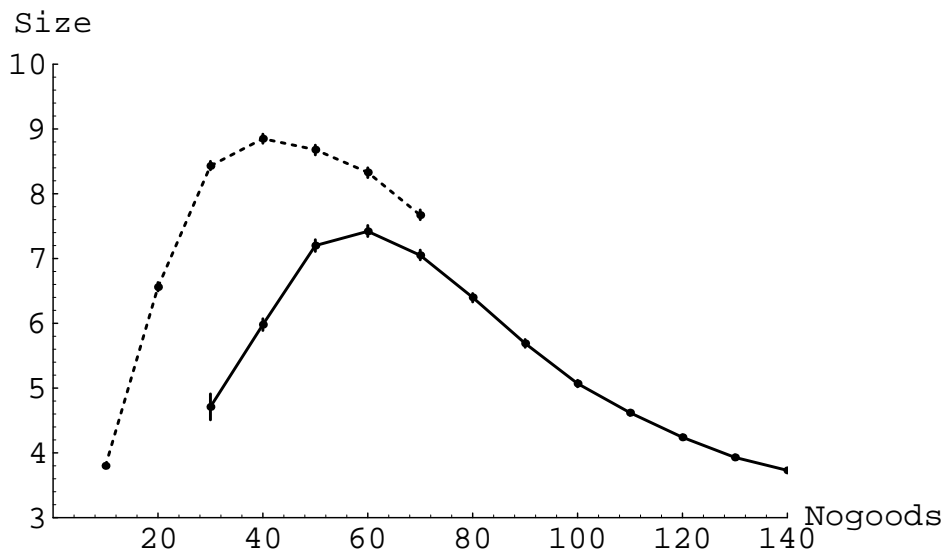

Figure 8: Mean size of smallest minimal unsolvable subproblem as a function of number of nogoods, for unsolvable problems generated using the "hill-climbing" (dotted line) and "generate-select" (solid line) methods. Each point is based on 1000 problems, except for the "generate-select" method at 30 nogoods, which is based on 100 problems. Error bars showing 95% confidence intervals are included.

The behavior of the minimal unsolvable subproblem size as a function of the number of constraints has a simple explanation. Unsolvable weakly constrained problems will generally need to concentrate most of the available constraints on a few variables in order to make all assignments inconsistent. This will tend to give one small minimal unsolvable subproblem. As more constraints are added, this concentration is no longer required and, since problems where most of the randomly selected constraints happen to be concentrated on a few variables are rare, we can expect more and larger minimal unsolvable subproblems.

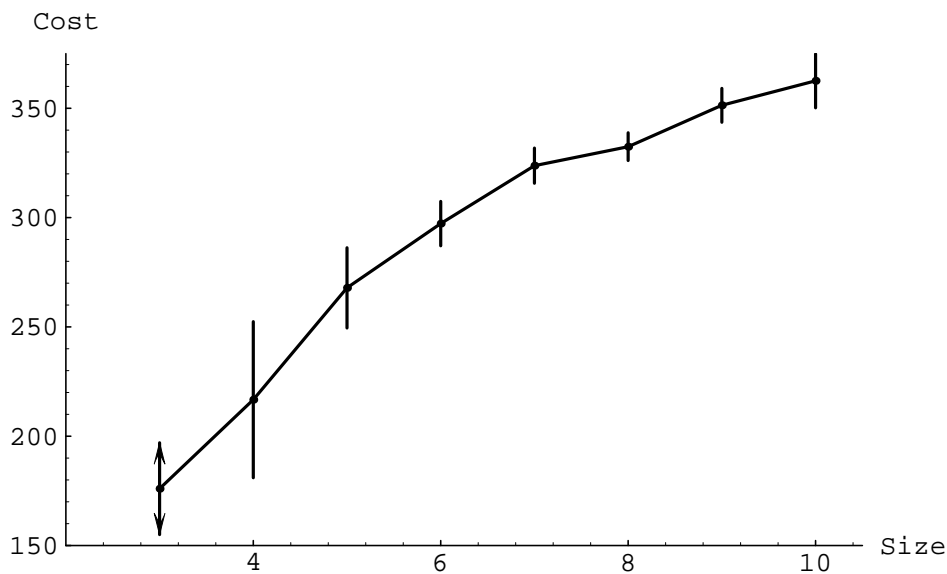

Figure 9: Mean solution cost as a function of size of smallest minimal unsolvable subproblem, for unsolvable problems with 60 nogoods generated using the "generate-select" method. Each point is the mean of the median solution costs, based on solving each problem 100 times, for the set of problems with the corresponding smallest minimal unsolvable subproblem size. The points are based on following numbers of problems for each smallest minimal unsolvable subproblem size, totaling 1000 problems: 3 − 1; 4 − 17; 5 − 71; 6 − 156; 7 − 253; 8 − 283; 9 − 165; 10 − 54. Error bars showing 95% confidence intervals are included, except for the single problem at size 3 for which confidence intervals cannot be calculated.

Finally, as more and more constraints are added, the increased pruning is equivalent to the notion that instantiating only a few variables is all that is required to find an inconsistency. This means we can expect a large number of small unsolvable subproblems. This qualitative description corresponds to what we observe in Figure 8.

Our observations of weakly constrained problems suggest that some search algorithms, such as dynamic backtracking, are able to rapidly focus in on one of the unsolvable subproblems and hence avoid the extensive thrashing, and high search cost, seen in other methods. In such cases, one would expect that the smaller the unsolvable subproblem, the easier it will be for the search to determine there are no solutions.

In order to examine the role of minimal unsolvable subproblem in search cost more closely, we plotted mean search cost versus size of smallest minimal unsolvable subproblem for unsolvable problems of 10 variables at each multiple of 10 nogoods from 30 to 140 nogoods. In every case, mean search cost increased as a function of size of smallest minimal unsolvable subproblem. Figure 9 shows an example of one of these plots, at the peak in solution cost for this class of problems, 60 nogoods. It makes sense that the smallest minimal unsolvable subproblem, being the easiest to detect, would play a significant role in search cost. However, the situation is surely more complicated than this, suggested by the fact that there is still variation among problems with the same size smallest minimal unsolvable subproblem. This could be due, for example, to one problem having several small minimal unsolvable subproblems, while another might have one minimal unsolvable subproblem, even smaller. Number and size of minimal unsolvable subproblems are both likely to play a role in search cost.

Number of minimal unsolvable subproblems does not seem to play as significant a role as size of smallest minimal unsolvable subproblem, but its effect can also be demonstrated. For the same sets of unsolvable problems as above, for each multiple of 10 nogoods from 80 to 140 nogoods, search cost correlates negatively with number of minimal unsolvable subproblems. However, for unsolvable problems with 30 to 70 nogoods, where variance in number of minimal unsolvable subproblems is lower (but variance in search cost is higher), there is no relationship between search cost and number of minimal unsolvable subproblems. Additional clarification of the role in search cost of both size and number of minimal unsolvable subproblems is left for further investigation. But size of smallest minimal unsolvable subproblem, which correlates strongly with search cost for (1) unsolvable problems taken as a whole (see Figures 2 and 8) and (2) unsolvable problems with a fixed number of nogoods over the full range of number of nogoods, appears to have the more primary effect.

This discussion of minimal unsolvable subproblems is also relevant to solvable problems: once a series of choices that precludes a solution is made during search, the remaining subproblem is now an unsolvable one. For example, in a 10-variable CSP, suppose values are given to the first two variables that are incompatible with all solutions to the problem. This means that in the context of these two assignments, the remaining eight variables constitute an unsolvable subproblem. The number of search steps required to determine that this subproblem is in fact unsolvable will be the cost added to the search before backtracking to the original two variables and trying a new assignment for one of them. Thus, the cost of identifying unproductive search choices for solvable problems is determined by how rapidly the associated unsolvable subproblem can be searched. As described above, when there are few constraints we can expect that such unsolvable subproblems will themselves have small minimal unsolvable subproblems and hence be easy to search with methods that are able to focus on such subproblems. While the unsolvable subproblems associated with incorrect variable choices in solvable problems may have a different structure, this argument suggests that changes in minimal unsolvable subproblems may explain the behavior of solvable problems with a fixed number of solutions as well. This could also explain observations of thrashing behavior for rare exceptionally hard solvable problems in the underconstrained region (Gent & Walsh, 1994a; Hogg & Williams, 1994); we would expect such problems to have a relatively large unsolvable subproblem to detect given the initial variable assignments. Finally, it would be interesting to study the behavior of local repair search methods for problems with a single solution to see if they also are affected by the change in minimal subproblem size.

## 6. Conclusions

We have presented evidence that the explanation of the easy-hard-easy pattern in solution cost based on a competition between changes in the number of solutions and pruning is insufficient to explain the phenomenon completely for sophisticated search methods. It does explain the overall pattern for problems not restricted by solvability or number of solutions. However, the explanation fails when the number of solutions is held constant and sophisticated search methods are used. In these cases the solution cost peak does not disappear as would be predicted. Alternatively, we can view this explanation as adequate for less sophisticated methods that are not able to readily focus in on unsolvable subproblems encountered during the search.

By considering relatively small search problems, we are able to exhaustively examine the properties of the search space. This allowed us to definitively demonstrate the importance for search behavior of an aspect of problem structure: the size of minimal unsolvable subproblems. Our approach contrasts with much work in this area that involves solving problems as large as feasible within reasonable time bounds. While the latter approach gives a better indication of the asymptotic behavior of the transition, it is not suitable for exhaustive evaluation of the nature of the search spaces encountered, nor for detailed analysis of aspects of individual problem structure.

We believe that detailed examination of the structure of combinatorial problems can yield information about why certain types of problems are difficult or easy. As a class, graph coloring or random CSPs are NP-complete, yet in practice many such problems are actually very easy. In addition, while theoretical work in this area has produced predictions that are asymptotically correct on average, the variance among individual problems in a predicted class is enormous. Increased understanding of the relationships between problem structure, problem solving algorithm, and solution cost is important to determining whether, and if so, how, we can determine prior to problem solving which problems are easy versus infeasibly hard. In contrast to previous theoretical studies that focus on the number of solutions, this work suggests that the size of minimal unsolvable subproblems is an alternate characteristic to study with the potential for producing a more precise characterization of the transition behavior and the nature of hard search problems.

## Acknowledgements

Much of this research was carried out while the first author was a summer intern at Xerox Palo Alto Research Center. This research was also partially supported by the National Science Foundation under Grant No. IRI-9321324 to Victor R. Lesser. Any opinions, findings, and conclusions or recommendations expressed in this material are those of the authors and do not necessarily reflect the views of the National Science Foundation.

*Conference on Artificial Intelligence*, pp. 304–310 Montréal, Québec, Canada.

Cheeseman, P., Kanefsky, B., & Taylor, W. (1991). Where the *really* hard problems are. In *Proceedings of the Twelfth International Joint Conference on Artificial Intelligence*, pp. 331–337 Sydney, Australia.

Crawford, J. M., & Auton, L. D. (1993). Experimental results on the cross-over point in satisfiability problems. In *Proceedings of the Eleventh National Conference on Artificial Intelligence*, pp. 21–27 Washington, DC, USA.

Gent, I. P., MacIntyre, E., Prosser, P., & Walsh, T. (1995). Scaling effects in the CSP phase transition. In Montanari, U., & Rossi, F. (Eds.), *Proc. of Principles and Practices of Constraint Programming PPCP95*, pp. 70–87. Springer-Verlag.

Gent, I. P., & Walsh, T. (1994a). Easy problems are sometimes hard. *Artificial Intelligence*, *70*, 335–345.

Gent, I. P., & Walsh, T. (1994b). The SAT phase transition. In Cohn, A. (Ed.), *Proceedings of the ECAI-94*, pp. 105–109. John Wiley and Sons.

Gent, I. P., & Walsh, T. (1996). The satisfiability constraint gap. *Artificial Intelligence*, *81*(1-2), 59–80.

Ginsberg, M. L. (1993). Dynamic backtracking. *Journal of Artificial Intelligence Research*, *1*, 25–46.

Hogg, T. (1996). Refining the phase transitions in combinatorial search. *Artificial Intelligence*, *81*, 127–154.

Hogg, T., & Williams, C. P. (1994). The hardest constraint problems: A double phase transition. *Artificial Intelligence*, *69*, 359–377.

Johnson, D. S., Aragon, C. R., McGeoch, L. A., & Schevon, C. (1991). Optimization by simulated annealing: An experimental evaluation; part II, Graph coloring and number partitioning. *Operations Research*, *39*(3), 378–406.

Kask, K., & Dechter, R. (1995). GSAT and local consistency. In *Proceedings of the Fourteenth International Joint Conference on Artificial Intelligence*, pp. 616–622 Montréal, Québec, Canada.

Kirkpatrick, S., & Selman, B. (1994). Critical behavior in the satisfiability of random boolean expressions. *Science*, *264*, 1297–1301.

Mitchell, D., Selman, B., & Levesque, H. (1992). Hard and easy distributions of SAT problems. In *Proceedings of the Tenth National Conference on Artificial Intelligence*, pp. 459–465 San Jose, CA, USA.

Prosser, P. (1996). An empirical study of phase transitions in binary constraint satisfaction problems. *Artificial Intelligence*, *81*, 81–109.